\documentclass{article}
\usepackage{ms}

\usepackage{times}
\usepackage{soul}
\usepackage{url}
\usepackage[hidelinks]{hyperref}
\usepackage[utf8]{inputenc}
\usepackage[small]{caption}
\usepackage{subcaption}
\usepackage{graphicx}
\usepackage{amsmath}
\usepackage{amsthm}
\usepackage{booktabs}
\usepackage{algorithm}
\usepackage{algorithmic}
\usepackage{amsfonts}
\usepackage{amssymb}
\usepackage{bm}
\usepackage{dsfont}
\usepackage{multirow}
\usepackage{multicol}
\usepackage{verbatim}
\usepackage{color}
\usepackage{float}

\newcommand{\tabincell}[2]{\begin{tabular}{@{}#1@{}}#2\end{tabular}}

\title{Towards Real-World Category-level Articulation Pose Estimation}% for Unseen Articulated Object with Unknown Kinematic Structure}

\author{
Liu Liu
\and
Han Xue\and
Wenqiang Xu\and
Haoyuan Fu\and
Cewu Lu
\affiliations
Shanghai JiaoTong University
\emails
\{liuliu1993, xiaoxiaoxh, vinjohn, simon-fuhaoyuan, lucewu\}@sjtu.edu.cn
}
\begin{document}

\maketitle

\begin{abstract}
%Articulated objects are pervasive in our daily life. Different from instance-level rigid object 6D pose estimation, real world articulated object understanding requires precise part-level 6D pose estimation for each category. In this paper, we aim to study the problem of estimating part-level 6D pose for multiple unseen articulated objects with unknown kinematic structures in a single RGB-D image. Specifically, we firstly propose a Convolutional Neural Network based articulated object detector to achieve multiple unseen category-level object detection. Then we develop a Point based encoder-decoder architecture to deal with part segmentation and part-level NOCS prediction for each detected instance separately. In addition, to cope with the issue of unknown kinematic structure for the unseen instance, we also propose a Joint Prediction module to estimate the information of unknown joint connecting rigid parts. To validate our method, we build our own articulated object model repository, on which we present a new Semi-Authentic MixEd Reality Technique that generates large amounts of fully annotated mixed reality data and also provide a real-world dataset with semi-automatic annotation pipeline. Experiments demonstrate that our method is able to robustly estimate part-level articulated object 6D pose in the real environments.

Human life is populated with articulated objects. Current Category-level Articulation Pose Estimation (CAPE) methods are studied under the single-instance setting with a fixed kinematic structure for each category. Considering these limitations, we reform this problem setting for real-world environments and suggest a CAPE-Real (CAPER) task setting. This setting allows varied kinematic structures within a semantic category, and multiple instances to co-exist in an observation of real world. To support this task, we build an articulated model repository ReArt-48 and present an efficient dataset generation pipeline, which contains Fast Articulated Object Modeling (FAOM) and Semi-Authentic MixEd Reality Technique (SAMERT). Accompanying the pipeline, we build a large-scale mixed reality dataset ReArtMix and a real world dataset ReArtVal. We also propose an effective framework ReArtNOCS that exploits RGB-D input to estimate part-level pose for multiple instances in a single forward pass. Extensive experiments demonstrate that the proposed ReArtNOCS can achieve good performance on both CAPER and CAPE settings. We believe it could serve as a strong baseline for future research on the CAPER task.

\end{abstract}

\section{Introduction}

Articulated objects are pervasive in our everyday life. Unlike rigid objects which can be regarded as a whole when moving in 3D space, articulated objects are usually composed of several rigid parts that are linked by different kinds of joints, e.g. revolute, prismatic, fixed, etc. In comparison with rigid objects, the diverse kinematic structures endow the articulated object higher Degree of Freedom (DoF), making the estimation of articulated object pose challenged.

Recently, the Category-level Articulated object Pose Estimation (CAPE) task has drawn increasing attention \cite{li2020category,yi2018deep}. Since the mechanism of estimating the articulation status from a single-view observation (e.g. RGB-D image, point cloud) can benefit downstream research and applications, such as scene understanding, robot manipulation, and VR/AR. However, currently, the task is generally studied under single-instance setting with synthetic point cloud, where the articulated object has a known and fixed kinematic structure for each category. Apparently, this assumption does not hold for many real-world cases. Specifically, (1) the synthetic object point cloud may have a domain gap for real-world applications, (2) daily objects may have different kinematic structures within a semantic category, e.g. drawers with different numbers of columns, (3) multiple objects may co-occur in a single observation. Given the gap between the current research direction and the real-world requirements, we extend the CAPE task by considering all the issues and reformulate the problem setting as CAPE-Real (short for \textbf{CAPER}). To support the CAPER task, we proposed a novel RGB-D based dataset \textbf{ReArtMix} which contains the objects from our proposed articulated model repository named \textbf{ReArt-48} for training the model and a baseline framework \textbf{ReArtNOCS} to address the CAPER task.

\begin{figure}[t]
    \centering
    \includegraphics[width=\linewidth]{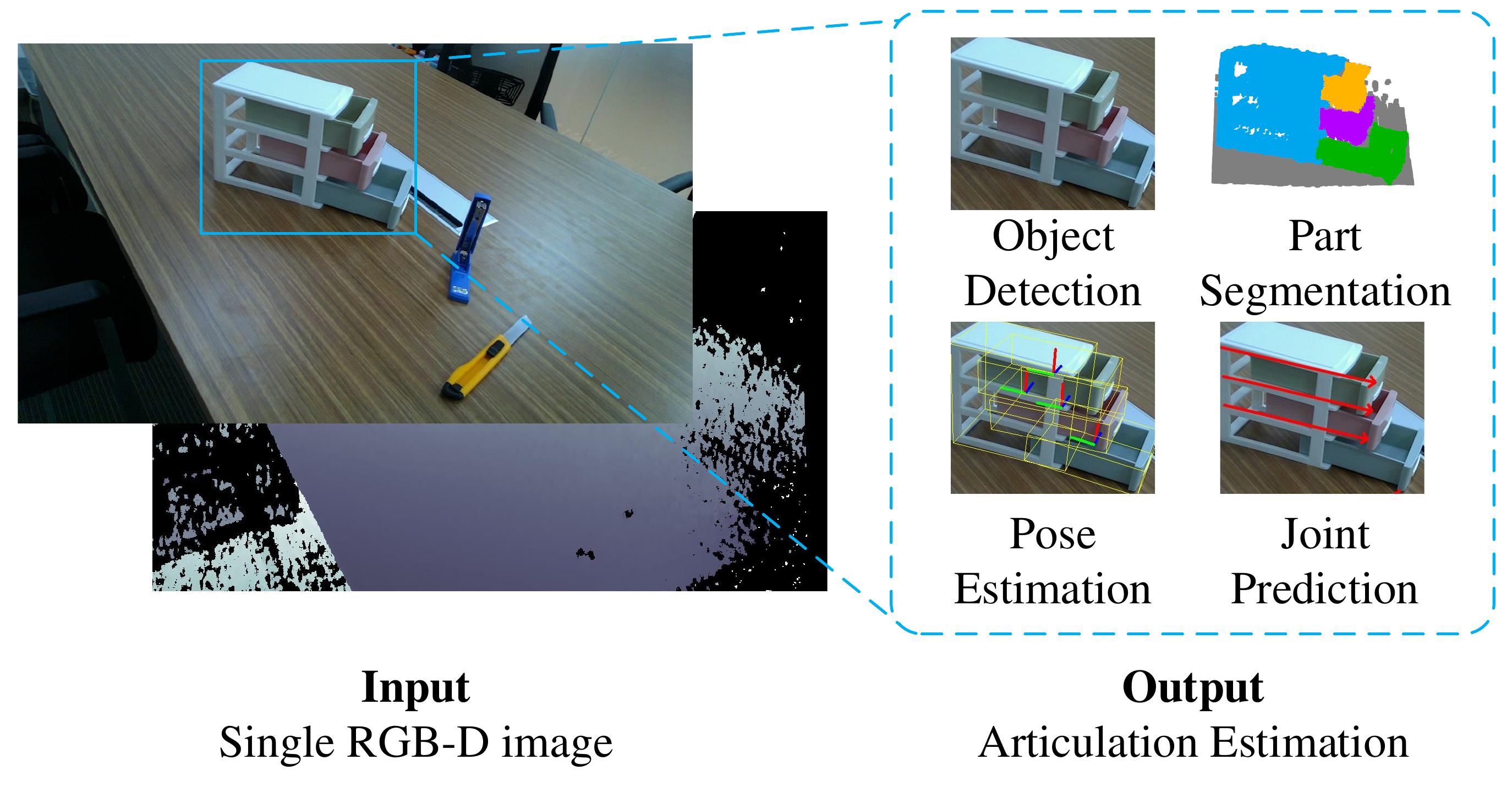}
    \caption{In real-world scenarios, the category-level articulated object pose estimation problem will face more challenges including varied kinematic structures within a semantic category, and multiple instances within a single RGB-D image. %\textbf{Real-World Articulated Object Pose Estimation and Joint Prediction.} We present a novel dataset tackling with articulated object understanding task, which is accompanied with a pipeline. Given a single RGB-D image captured in the real world, we aim to achieve articulated object detection and then segment each rigid part, estimate part-level 6D pose and predict joint information.
    }
    \label{fig:abstract}
\end{figure}

When constructing the dataset ReArtMix, two major challenges exist: First, there is no suitable public model repository for real articulated objects, current popular articulated model repositories either contain only synthetic models \cite{xiang2020sapien,wang2019shape2motion} or support only instance-level task \cite{martin2019rbo}. Second, collecting RGB-D training data with articulated pose annotations are cost-prohibitive. Therefore, we take a synthetic path to create the dataset. We first build a \textbf{Re}al-world \textbf{Art}iculated model repository named \textbf{ReArt-48} which contains 48 different scanned models under 5 categories. The part segmentation and joint properties of the scanned models are annotated through a \textbf{F}ast \textbf{A}rticulated \textbf{O}bject \textbf{M}odeling (\textbf{FAOM}) pipeline. Once obtaining the annotated object models, we composite them with real-world background RGB-D images in a physically plausible manner and automatically generate a large-scale mixed reality dataset \textbf{ReArtMix} along with the annotations required (e.g. object part segmentation, part pose, joint properties, etc.) by our proposed \textbf{S}emi-\textbf{A}uthentic \textbf{M}ix\textbf{E}d \textbf{R}eality \textbf{T}echnique (\textbf{SAMERT}). To prove ReArtMix can effectively reduce domain gap when transferring to real-world scenarios, we also build a fully real-world dataset \textbf{ReArtVal} for validation. The quantitative results are reported in Table \ref{tab:comparison exp}. In comparison with annotating real-world images totally by a human ($\sim$ 2 min/image), the FAOM-SAMERT pipeline can save a proliferation of  human labors for image capturing and annotation ($\sim$ 0.2 sec/image).

Accompanying the dataset ReArtMix, we propose a learning framework named \textbf{Re}al-world \textbf{Art}iculation \textbf{NOCS} (\textbf{ReArtNOCS}), as it is inspired by Normalized Object Coordinate Space (NOCS) \cite{wang2019normalized}. ReArtNOCS can utilize both RGB and depth information, handle multiple instances in a single forward pass. It consists of an RGB-D based object detector, a point-cloud based part predictor, and a joint property predictor that can adapt to varied kinematic structures for detected instances. To evaluate the framework, we test our method on the proposed dataset ReArtMix with Average Precision on rotation error, translation error, and 3D bounding box IoU as metrics, and also test on CAPE setting of samples rendered with PartNet-Mobility \cite{xiang2020sapien}.

Our contributions can be summarized in three folds: (1) we bring the previously proposed CAPE problem towards real-world setting as CAPER problem, which is more realistic and complicated as we consider multiple unseen objects with varied kinematic structures in a semantic category. (2) We collect a real-world articulated model repository ReArt-48 and propose a FAOM-SAMERT pipeline to make the preparation of the real-world-like training dataset feasible. (3) We propose an effective framework ReArtNOCS to address the CAPER problem, which could serve as a strong baseline for the task. 

\section{Related Work}

\textbf{Category-level 6D pose estimation} Instance-level pose estimation aims to predict objects' 3D rotation and translation given 3D object models \cite{kehl2017ssd,wang2019densefusion}. On the contrary, the goal of category-level pose estimation is to predict an input instance's pose and location relative to category-specific representation. The first proposed method is to predict 3D-3D per-pixel correspondences between observations and canonical coordinates using a normalized space for category-level representation \cite{wang2019normalized}. Besides, \cite{manhardt2020cps} present to directly optimize the predicted rotation, translation, and scale simultaneously in a single monocular image. In addition to the static image-based estimator, \cite{wang20196} propose the first category-level pose tracker, which adopt geometric or semantic keypoints to estimate the interframe motion. Among these methods, they mainly focus on rigid pose estimation while our task aims to estimate articulation pose in the real world.

%On the contrary, the goal of category-level pose estimation is to infer an unseen instance's location, size and pose relative to a canonical object representation. One of the typical approaches is proposed by NOCS that normalized the scales and universally align the orientations for objects in a given category \cite{wang2019normalized}. So, this work can handle unseen objects in a given category by predicting 3D-3D correspondences. Based on NOCS, another similar work focuses on pose estimation in single monocular data by directly optimizing over 3D rotation, translation, scale and shape simultaneously using a novel 3D pose point cloud loss \cite{manhardt2020cps}. 

\textbf{Articulated object pose estimation} Articulation estimation has been studied for decades, where previous works can be broadly categorized into interaction-based, video-based, and single-view-based. Interaction-based methods are mostly studied in the Robotics community, which adopt feedback from manipulation actions to characterize a distribution of articulation models \cite{hausman2015active}. The interaction-based approaches require some ways to interact with the object, which limits the applicability. On the other hand, a video recorded with a moving articulated object can be a good source to estimate its motion property in an image sequence \cite{liu2020nothing}. With a simplified setting, instead of a clip of video, in which Yi et al. take a pair of unsegmented shape representations as input to predict correspondences, 3D deformation flow and part-level segmentation \cite{yi2019deep}. With the development of deep learning techniques, the single-view-based CAPE setting is becoming possible. A-NCSH \cite{li2020category}, as an extension of NOCS, is developed for single articulated object pose estimation. However, it holds an obvious limitation that requires fixed kinematic structure as prior information for each input object while our CAPER setting allows multiple instances and various kinematic structures.

\begin{figure}[t]
    \centering
    \includegraphics[width=\linewidth]{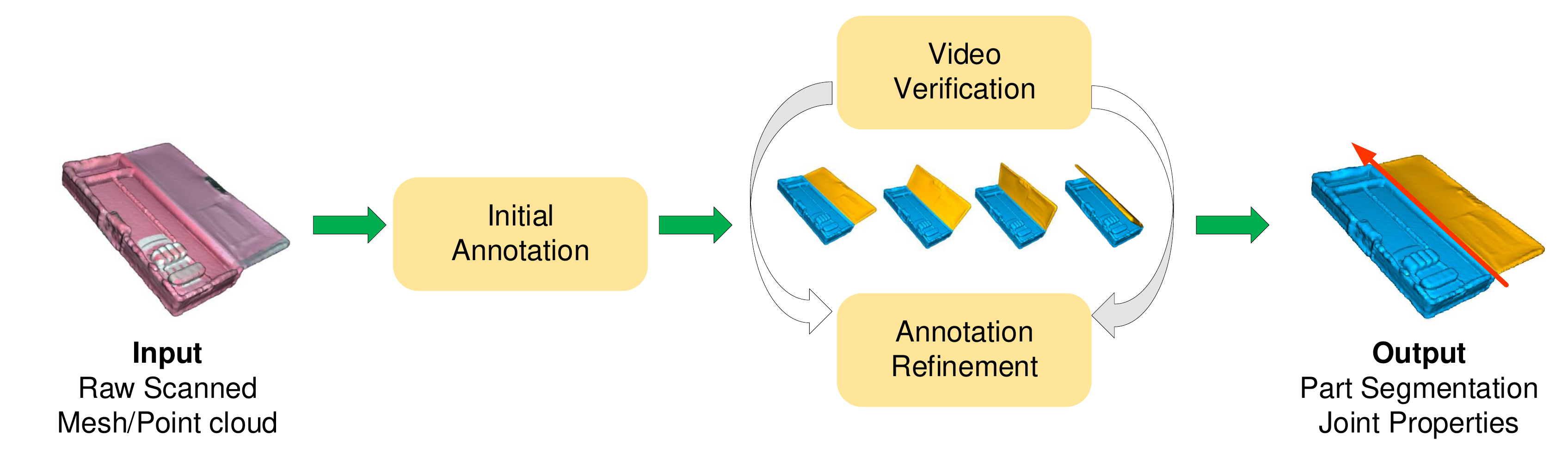}
    \caption{\textbf{Fast Articulated Object Modeling} pipeline. We use an iterative method to refine the initial articulation annotation.}
    \label{fig:faom}
    \vspace{-0.5cm}
\end{figure}

\section{Problem Statement}
As mentioned earlier, the CAPER problem setting advances the CAPE setting in three aspects: object shape and texture are realistic, multi-instance in an image, and objects of the same semantic category have different kinematic structures.

In CAPER problem, given a single RGB-D image $I$ as input, a CAPER model will firstly learn to detect $N$ objects with bounding box $\mathcal{O} = \{(u_1^i,v_1^i,u_2^i,v_2^i)\}_{i=1}^N$ with left top $(u_1^i,v_1^i)$ and right bottom $(u_2^i,v_2^i)$ coordinates as well as their corresponding categories $\mathcal{C}=\{c^i\}_{i=1}^N$. In dealing with each detected instance, we project the instance patch of $i^\text{th}$ instance to local point cloud $\mathcal{P}^i=\{(x^i_j, y^i_j, z^i_j, r^i_j, g^i_j, b^i_j)\}_{j=1}^M$ which has $M$ points and predict: (1) segmentation for unknown number of parts $\mathcal{S}=\{s_j\}_{j=1}^M$, where $s_j$ is the index of one of the maximum $K$ parts; (2) part-level NOCS map $\mathcal{P'}$ describing canonical representation $\mathcal{P'}=\{\textbf{p}'_j=(x'_j, y'_j, z'_j) \in \mathbb{R}^3\}_{j=1}^M$ (we use symbol $'$ to define the coordinate in NOCS space); (3) joint properties that describes joint type $\delta_j$, location $\textbf{q}_j$ and axis $\textbf{u}_j$. Finally, given these prediction results, we recover the 3D rotation $R^k$ and 3D translation $\textbf{t}^k$ for $k^\text{th}$ part. 

%To unify the representation of rigid and articulated objects, instead of adopting the object center as the location as classic 6DOF estimation problem do \cite{}, we regard the rigid object as a part of the world, and the joint between the rigid object and the world is defined to be fixed. Formally, the location of an object is translated from the center to the bottom of the object along the backward of the orientation. In this sense, the rigid object can be regarded as a type of special articulated object.

\section{Datasets}\label{dataset}
As there exists no dataset to fully support the CAPER task, we construct the ReArtMix dataset by taking a mixed reality approach to reduce the human labor for annotation. Firstly, we collect scanned models of common real objects with varied kinematic structures and categories, and annotate part segmentation along with joint properties of the objects using the proposed FAOM pipeline (Sec. \ref{sec:FAOM}). Then, we composite these articulated models with real-world background  RGB-D images to obtain training samples with full annotations (Sec. \ref{sec:SAMERT}). Last but not the least, we build up a real dataset for validation (Sec. \ref{sec:real_data}).

\subsection{FAOM, Fast Articulated Object Modeling}\label{sec:FAOM}

%In this section, we first introduce the object repository and then describe the semi-automatic articulation modeling (FAOM) process. 

\paragraph{Model repository and annotation} We scanned 48 hand-scale objects from 5 common categories (such as box and stapler) in our daily life with EinScan Pro 2020\footnote{\url{https://www.einscan.com}}. To convert these scanned models to articulated models, instead of adopting the traditional 3D reverse engineering software\footnote{\url{https://www.3dsystems.com/software/geomagic-design-x}} which is tedious and requires expertise, we propose a fast modeling method FAOM illustrated in Fig. \ref{fig:faom}. In FAOM, we achieve video verification by animating the initial annotated joint properties and part segmentation that helps a lot in annotation refinement. By the iterative process, raw scanned models can be easily annotated with much less labor. Please refer to the supplementary file for more details about ReArt-48.

\paragraph{Comparison with other model repositories} As shown in Table \ref{tab:model_comparison}, our object models have full features required for the CAPER task. Compared to other real model repositories, ours are more than twice the object number. %Therefore, rendering with our articulated models could own much more details which provide powerful information for us to estimate 6D pose and joint state in the real world.

\begin{table}[tbh]
%\scriptsize
\centering
\resizebox{\linewidth}{!}{
\begin{tabular}{l|cccc}
\hline
Model repository & Articulation & Real & Category-level & \tabincell{c}{Object \\ Num}\\
\hline
ShapeNet\cite{chang2015shapenet} & & & \checkmark & $>$50K \\
YCB\cite{calli2015ycb} &  & \checkmark & & 21 \\
LineMod\cite{2012Multimodal} &  & \checkmark & & 15\\
Shape2Motion\cite{wang2019shape2motion} & \checkmark & & \checkmark & 2,440  \\
PartNet-Mobility\cite{xiang2020sapien} & \checkmark & & \checkmark & 2,346  \\
RBO\cite{martin2019rbo} & \checkmark & \checkmark &  & 14  \\
\hline
Ours & \checkmark & \checkmark & \checkmark & 48\\
\hline    
\end{tabular}}
\caption{Comparison with other popular model repositories. }
\label{tab:model_comparison}
\vspace{-0.5cm}
\end{table}

\begin{comment}
\begin{table}[tbh]
%\scriptsize
\centering
\caption{Comparison with other popular model repositories. Note that our models are captured from real world with much larger number of triangular patches than some other synthetic models, which indicates that finer details of objects are considered in our dataset.}
\resizebox{\linewidth}{!}{
\begin{tabular}{l|cccc}
\hline
Benchmark & \tabincell{c}{Object \\ Type} & \tabincell{c}{Object \\ Num} & Real? & \tabincell{c}{Kinematic \\ Structure}\\
\hline
ShapeNet & rigid & $>$50K & No & -\\
YCB & rigid & 21 & Yes & -\\
LineMod & rigid & 15 & Yes & - \\
Shape2Motion & articulated & 2,440 & No & single \\
SAPIEN & articulated & 2,346 & No & single \\
RBO & articulated & 14 & Yes & single \\
\hline
Ours & articulated & 48 & Yes & multiple \\
\hline    
\end{tabular}}
\label{tab:model_comparison}
\end{table}
\end{comment}

\begin{figure}[t]
    \centering
    \includegraphics[width=0.9\linewidth]{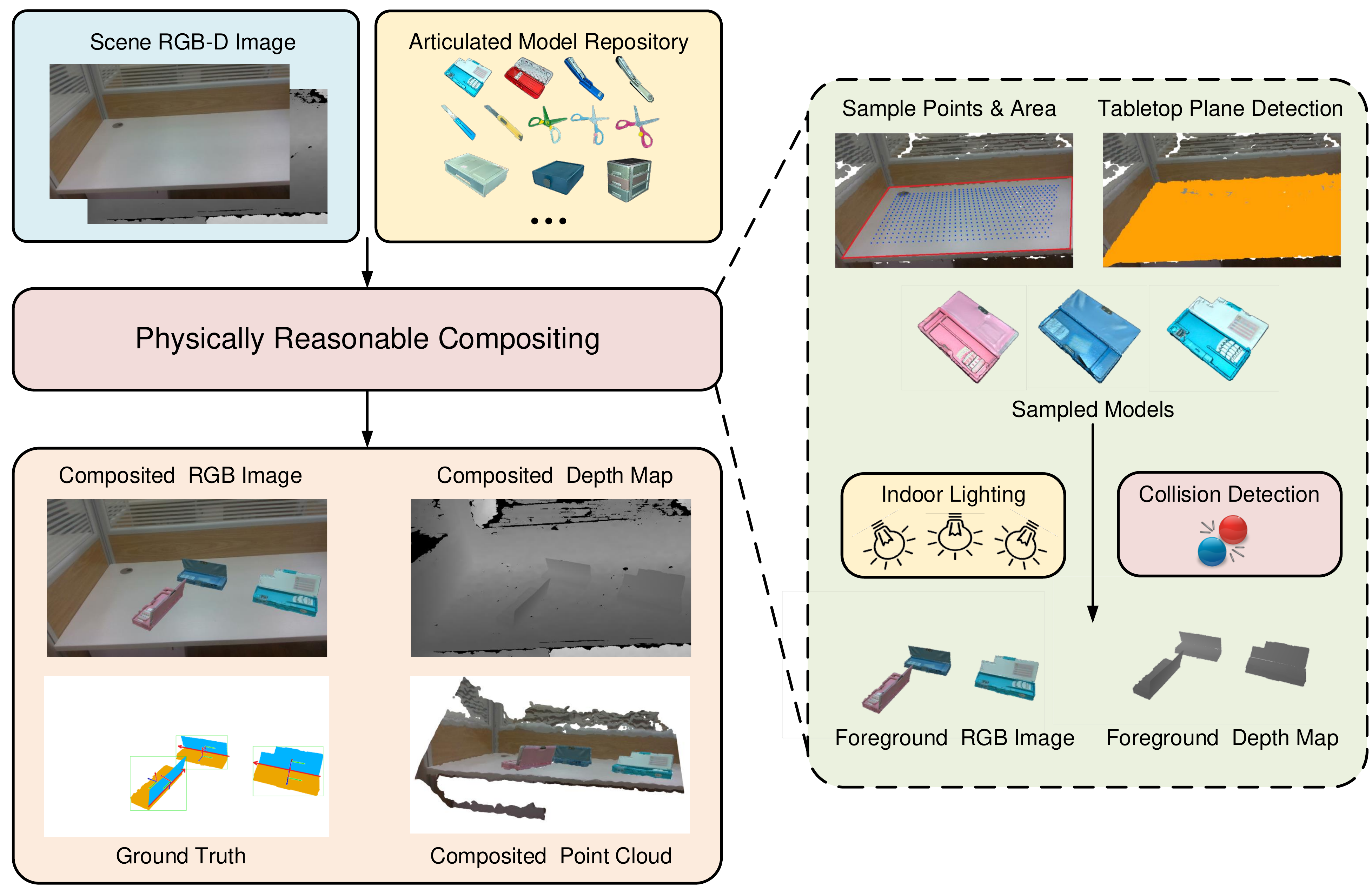}
    \caption{Our \textbf{Semi-Authentic MixEd Reality Technique (SAMERT)}. We combine articulated models scanned from real objects with RGB-D images captured from real scenes to generate real-looking composited RGB-D images. Plane detection and collision detection are performed to make the synthetic data physically reasonable.}
    \label{fig:synthetic_pipeline}
    \vspace{-0.5cm}
\end{figure}

\begin{figure*}[ht]
    \centering
    \includegraphics[width=\linewidth]{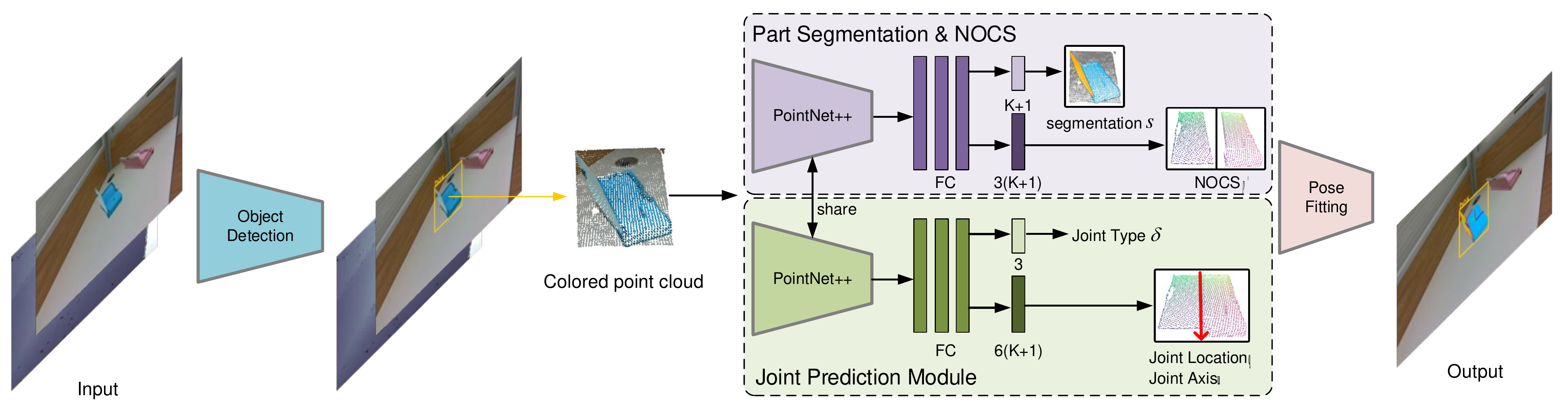}
    \caption{\textbf{ReArtNOCS} for real world articulated object pose estimation. There are three modules: object detection, part segmentation \& NOCS and joint prediction. In our framework, we detect each articulated object from RGB-D image and predict each instance's part-level NOCS with corresponding joint properties.}
    \label{fig:method}
    \vspace{-0.5cm}
\end{figure*}

\subsection{SAMERT, ReArtMix Dataset Generation}\label{sec:SAMERT}

The training data for the CAPER task requires realistic visual quality and high-precision annotations. However, the accurate annotation for the part-level pose is time-consuming, which is prohibitive for large-scale dataset preparation. To address this, we propose a novel \textbf{Semi-Authentic MixEd Reality Technique (SAMERT)} based on Unity Engine\footnote{\url{https://unity.com}} to automatically generate such training dataset with articulated models and pre-collected real-world RGB-D background images. These background images are snapshots from 20 different tabletop scenes with $\sim 200$ viewpoints per scene. Given the real-scanned object models and background images, we present a \textsl{physically reasonable compositing} strategy to render RGB-D images with full annotations. Firstly we randomly place the articulated objects with random scale and joint state onto the desktop plane of the background image in 3D camera space, which is predicted by PlaneRCNN \cite{liu2019planercnn} and RANSAC algorithm. Part-level object pose and joint states are tracked and recorded, as well as the collision status among objects. The physics engine will guarantee the physically plausible requirements, that is, no floating objects and intersections between objects. We also randomly generate multiple directional lights with different orientations and colors to imitate indoor lighting. Please refer to the supplementary file for details of the implementation.

%we randomize the orientation of multiple directional lights applying in foreground objects. Next, given the depth maps of foreground objects and background scenes, we adopt PlaneRCNN\cite{liu2019planercnn} and RANSAC algorithm to obtain the predicted plane from a background scene, e.g. desktop, and randomly sample placement points as object locations in the scene. During the location and pose generation of multiple instances, collision detection are performed in these steps to satisfy the physical requirements. The pipeline of our synthetic data generation is shown in \ref{fig:synthetic_pipeline}. More details can be found in supplementary materials.

%Given the depth maps of foreground objects and background scenes, we satisfy following requirements to improve realism of the composited RGB-D images: (1) All the objects are located exactly onto the desktop plane in 3D space as they are seen in the real world. (2) Different instances of objects do not overlap with each other in 3D space and background scene e.g. the wall. Under these requirements, we adopt PlaneRCNN\cite{liu2019planercnn} and RANSAC to get plane detection results for background scenes and the sample placement points are randomly generated as object locations. During the location and pose generation of multiple instances, collision detection are performed in these steps to satisfy the physical requirements.

With the SAMERT process, we generate 100K RGB-D images, of which 90K are set aside for training and 10K for validation. Among these images, 37 articulated models in ReArt-48 are used to generate training images while the rest of 11 objects are selected as unseen for validation. With the real scanned models and real background scenes, our synthetic semi-authentic data could drastically reduce the gap between virtuality and reality, which can be quantitatively proved with the real validation set described next.

%Please refer to the supplementary file for the specification of the data split.

\subsection{ReArtVal, Real Data Acquisition for Validation}\label{sec:real_data}
To validate the performance of our method in the real world, we also build a fully real dataset in the form of video sequences. For each category, we capture over 6K RGB-D frames in 6 real-world tabletop scenes using RealSense D435i camera\footnote{\url{https://www.intelrealsense.com/depth-camera-d435/}}. In terms of data annotation, we propose a semi-automatic part-level 6D pose annotation pipeline referred by LabelFusion \cite{marion2018label}, in which there are two steps in annotating the video: (1) Manual Process. The initial frame RGB-D image of the video is annotated manually by the 3-click method for coarse alignment and then we apply ICP for the refinement. (2) Given the precise annotations for an initial frame, we adopt colorized point cloud for RGB-D registration between the initial frame and any other in the input video. Finally, we capture over 6K RGB-D frames with full annotations (bounding box, part segmentation, part-level 6D pose, joint properties) to build our real-world dataset. Please refer to the supplementary file for the details of the annotation process. 

\section{Method}

We propose a framework ReArtNOCS to address the CAPER task. The key challenge for ReArtNOCS is to handle multiple instances within a single image and varied kinematic structures (unknown joint properties) within a semantic category. The multi-instance problem can be tackled by object detection (Sec. \ref{sec:object detection}). And a joint prediction module is designed to handle the detected instance with varied kinematic structures (Sec. \ref{sec:articulation pose}). Accompanying with predicted NOCS map and joint properties, we could recover per part pose. The training is formulated as a multi-task learning problem (Sec. \ref{sec:multi task loss}). The overall pipeline is displayed in Fig. \ref{fig:method}.

%In dealing with each detected instance with unknown kinematic structure, we predict for each pixel/point over the articulated parts the following: (1) segmentation map for unknown number of parts; (2) part-level NOCS map describing canonical representation. (3) joint information that describes the kinematic structure including joint type (fixed, prismatic, revolute) and joint axis in in canonical space.

%To address the unseen articulated object 6D pose estimation with unknown kinematic structure, we design an effective framework containing articulated object detector, part-level segmentation and NOCS prediction, joint information estimation. The framework is shown in Fig. \ref{fig:method}.

\subsection{Object Detection}\label{sec:object detection}

Accurate detection performance is a high priority for our method. Here we adopt an anchor-based object detector RetinaNet \cite{lin2017focal} for dense detection with RGB-D image, where the input contains 6 channels (3 for RGB and 3 for XYZ). Since unseen articulated objects usually hold various appearances in texture, we adopt texture augmentation\cite{borrego2018applying} to improve the appearance diversity of the training set. Specifically, we randomly generate flat colors, gradients of colors, chess patterns, and Perlin noise as object textures.

\subsection{Articulation Pose Estimation}\label{sec:articulation pose}

\paragraph{Part Segmentation and NOCS Prediction}
For each detected bounding box $\mathcal{O}^i$, we crop the corresponding region in the RGB-D images and transform the patch to colored point cloud $\mathcal{P}^i$. Next, the local point cloud is processed by a PointNet++ \cite{qi2017pointnet} architecture for feature extraction. At the end of PointNet++, we build two parallel branches with $K+1$ and $3(K+1)$ channels for part segmentation and part-level NOCS prediction, where $K$ is the maximum number of rigid parts in our dataset and $1$ indicates the background. The part-level NOCS map is defined for each separate rigid part rather than the whole object, in which we define the rest state for every part and normalized the shape of the rest state. Finally, in ReArtNOCS, we could predict part segmentation label $s_j$ and part-level NOCS coordinate $\textbf{p}'_j$ on $j^\text{th}$ point.

%Finally, after segmenting each 3D point $p_j$ with predicted label $s_j$, we could obtain the final part-level NOCS coordinate $q_j$ per point by:

%\begin{equation}
%q_j = \mathop{\arg\max}_{k} \{q_j^k \times s_j^k | k=1,2,...,K\} 
%\end{equation}

\paragraph{Joint Prediction}

Current methods such as A-NCSH for the CAPE setting requires fixed kinematic structure as prior knowledge \cite{li2020category} . In our method, we propose a joint prediction module to handle varied kinematic structures in one semantic category. Here we assume that all the parts and joints have one-to-one correspondence. Therefore we could predict joint properties for each corresponding segmented part and the adaptive kinematic structure could be dynamically solved in ReArtNOCS.  We summarize three different joint types, including fixed, prismatic and revolute.
%Based on this, we could predict joint information corresponding to each segmented part. Since the number of parts are unknown during part segmentation process, the unknown kinematic structure could be dynamically predicted in our framework. Therefore, our method could handle the "unseen and unknown structured" articulated objects.

In our joint prediction module, we aim to predict three types of information for each kinematic joint: joint type $\delta^k$ (also known as kinematic way), joint location $\textbf{q}^k$ and joint axis $\textbf{u}^k$. As we assign every part to its corresponding joint, we could densely predict joint information in our framework. Specifically, we introduce two new branches into the end of PointNet++  with $3$ and $6(K+1)$ channels, in which $3$ indicates that the joint type prediction aims to classify each part into three kinematic way $\delta{_j}$ and $6(K+1)$ channels are used to regress the joint location $\textbf{q}_j$ and joint axis $\textbf{u}_j$ with non-class agnostic way. For the final joint location $\textbf{q}^k$, we follow the voting scheme from A-NCSH. For joint axis $\textbf{u}^k$, we average the $\textbf{u}_j^k$ on points from part $k$:

\begin{equation}
\textbf{u}^k = \frac{\sum_{j=1}^M \textbf{u}_j^k \mathds{1}(s_j=k)}{\sum_{j=1}^M \mathds{1}(s_j=k)}
\end{equation}

\subsection{Multi-task Loss Function}\label{sec:multi task loss}

For articulated object detection task, we use focal loss \cite{lin2017focal} and smoothL1 loss for bounding box classification and regression $\mathcal{L}_{det}=\mathcal{L}_{FL}+\mathcal{L}_{smoothL1}$. In terms of PointNet++ based network, the total loss for articulated object estimation $\mathcal{L}_{ReArt}$ is:

\begin{equation}
\begin{split}
\mathcal{L}_{ReArt} &= \lambda_{1}\mathcal{L}_{seg} + \lambda_{2}\mathcal{L}_{nocs} \\ 
&= + \lambda_{3}\mathcal{L}_{loc} +  \lambda_{4}\mathcal{L}_{ax} + \lambda_{5}\mathcal{L}_{type}
\end{split}
\end{equation}

Specifically, we use the cross-entropy loss for part segmentation task $\mathcal{L}_{seg}$. Then L2 is adopted as NOCS map loss $\mathcal{L}_{nocs}$, joint location $\mathcal{L}_{loc}$ and joint axis $\mathcal{L}_{ax}$ prediction tasks. Finally, we use IoU loss for joint type classification $\mathcal{L}_{type}$:

\begin{equation}
\begin{aligned}
& \mathcal{L}_{seg} = \sum\nolimits_{j=1}^{M} CE(s_j, s_j^*) \\
& \mathcal{L}_{nocs} = \sum\nolimits_{j=1}^M \mathds{1}(s_j^* > 0){\left\| \textbf{p}^\prime_j - \textbf{p}^{\prime*}_j \right\|_2} \\
& \mathcal{L}_{loc} = \sum\nolimits_{j=1}^M \mathds{1}(s_j^* > 0){\left\| \textbf{q}_j - \textbf{q}_j^* \right\|_2} \\
& \mathcal{L}_{ax} = \sum\nolimits_{j=1}^M \mathds{1}(s_j^* > 0){\left\| \textbf{u}_j - \textbf{u}_j^* \right\|_2} \\
& \mathcal{L}_{type} = \sum\nolimits_{j=1}^M \mathds{1}(s_j^* > 0)(1 - IoU(\delta{_j}, \delta{_j}^*)) \\
\end{aligned}
\end{equation}

\begin{comment}
\begin{equation}
\begin{split}
\mathcal{L}_{ReArt} &= \lambda_{1}\mathcal{L}_{seg} + \lambda_{2}\mathcal{L}_{nocs} \\ 
&= + \lambda_{3}\mathcal{L}_{loc} +  \lambda_{4}\mathcal{L}_{ax} + \lambda_{5}\mathcal{L}_{type} \\
 &= \sum_{j=1}^M \lambda_{1}CE(s_j, s_j^*) + \lambda_{2}[s_j^* > 0]{\left\| \textbf{q}_j - \textbf{q}_j^* \right\|_2} \\
 % \sqrt{(\textbf{q}_j - \textbf{q}_j^*)^2} \\
 & + \lambda_{3}[s_j^* > 0]{\left\| \mu{_j} - \mu{_j}^* \right\|_2} \\
 & + \lambda_{4}[s_j^* > 0]{\left\| \Phi{_j} - \Phi{_j}^* \right\|_2} \\
 % \sqrt{(\Phi{_j} - \Phi{_j}^*)^2} \\
 & + \lambda_{5}[s_j^* > 0](1 - IoU(\delta{_j}, \delta{_j}^*)) \\
\end{split}
\end{equation}
\end{comment}

where the five multiplication factors $\lambda_{1}$, $\lambda_{2}$, $\lambda_{3}$, $\lambda_{4}$, $\lambda_{5}$ are set to be 1, 10, 1, 0.5, 1. $s_j^*$, $\textbf{p}^{\prime*}_j$, $\textbf{q}_j^*$, $\delta{_j}^*$ and $\textbf{u}_j^*$ denote the ground truth of part labels, part-level NOCS map, joint location, joint axis and joint type respectively. $\mathds{1}(s_j^* > 0)$ indicates the loss is only accounted when the part is foreground. Finally, with predicted part segmentation, NOCS map and joint properties, we follow the pose optimization algorithm with kinematic constrains \cite{li2020category} to recover the 6D pose for each rigid part.

\section{Experiments}

\begin{figure}[!tb]
    \centering
    \begin{subfigure}[t]{0.3\linewidth}
    \includegraphics[width=\linewidth]{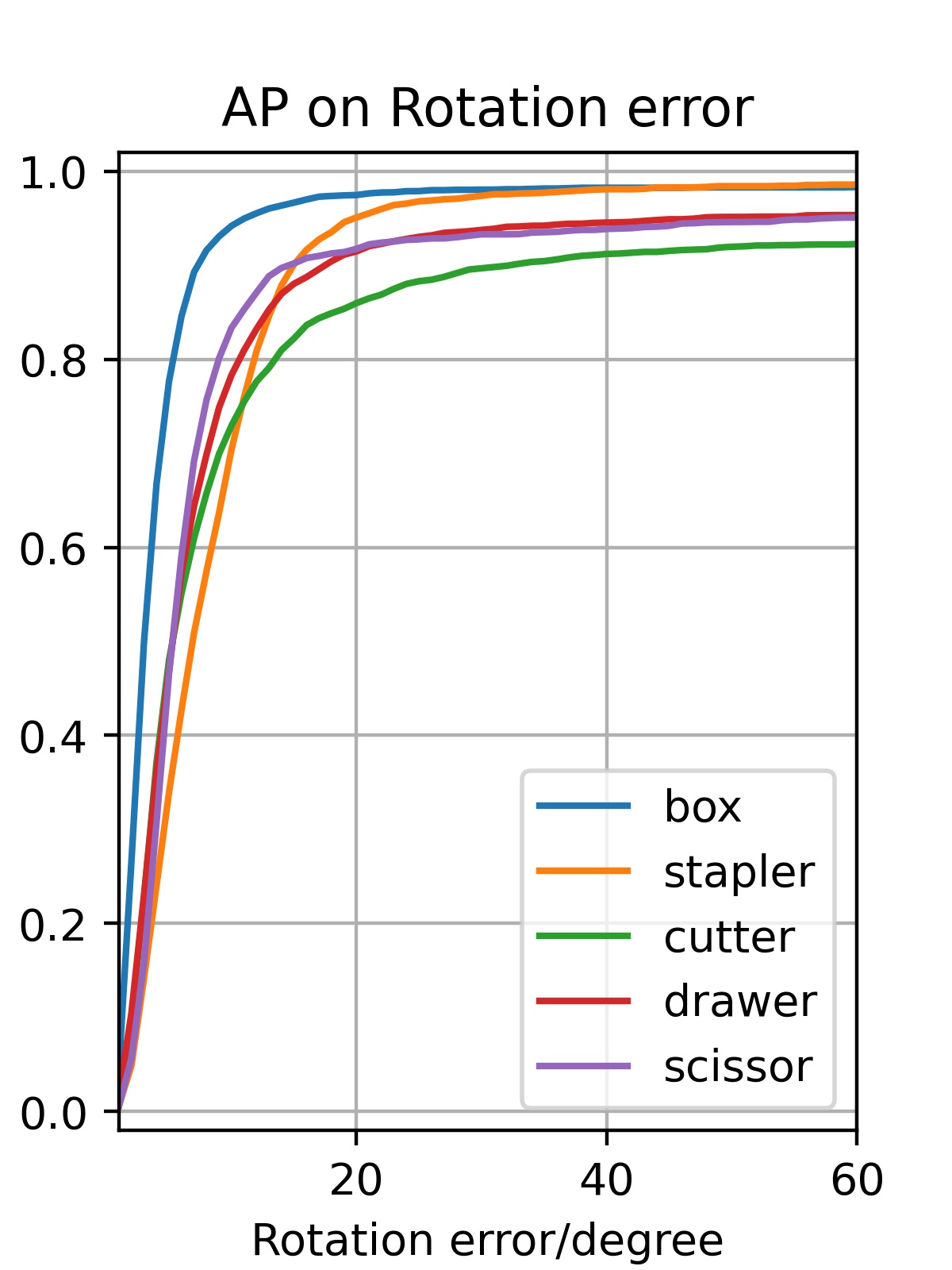}
    \end{subfigure}
    \begin{subfigure}[t]{0.3\linewidth}
    \includegraphics[width=\linewidth]{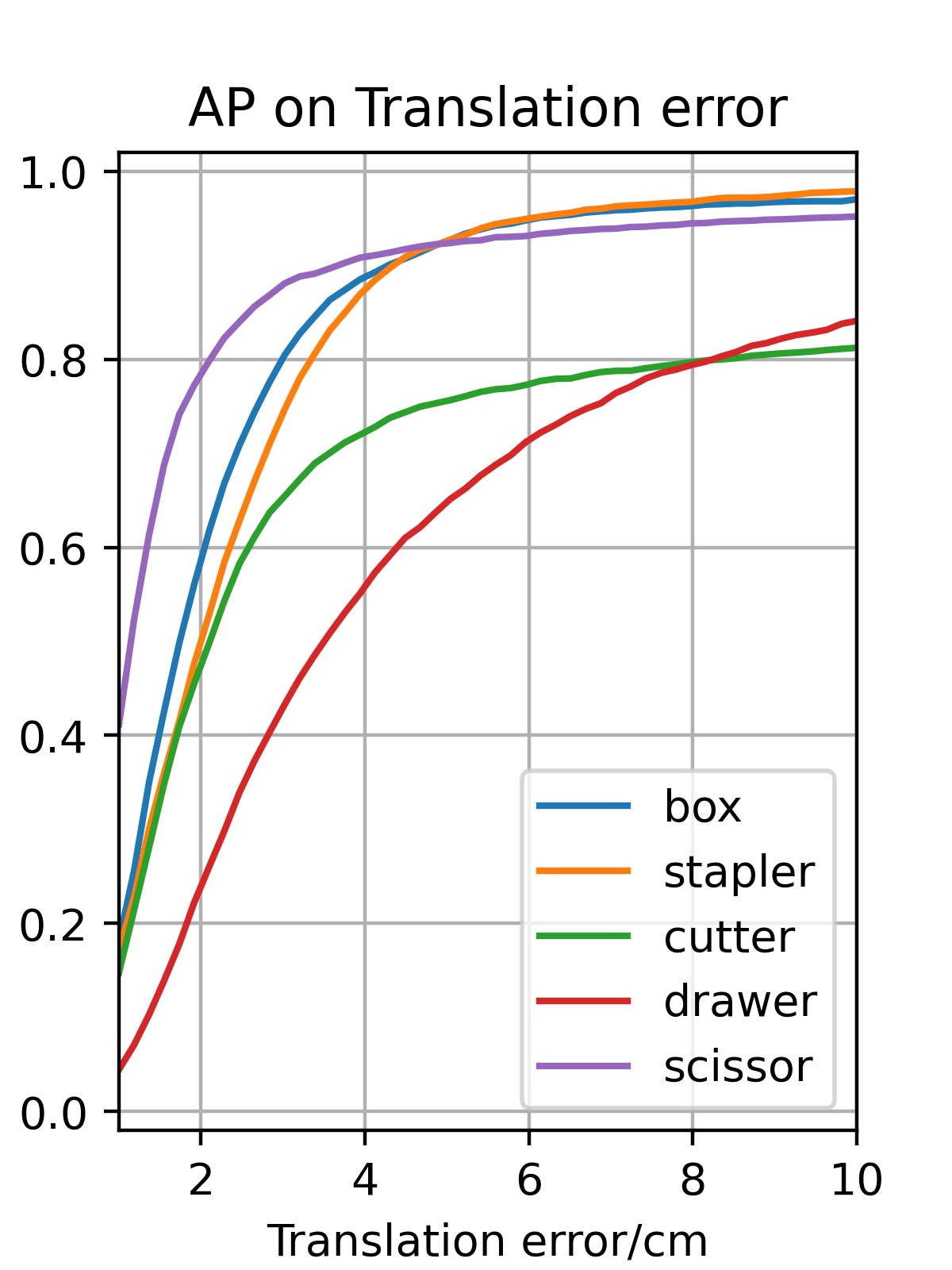}
    \end{subfigure}
    \begin{subfigure}[t]{0.3\linewidth}
    \includegraphics[width=\linewidth]{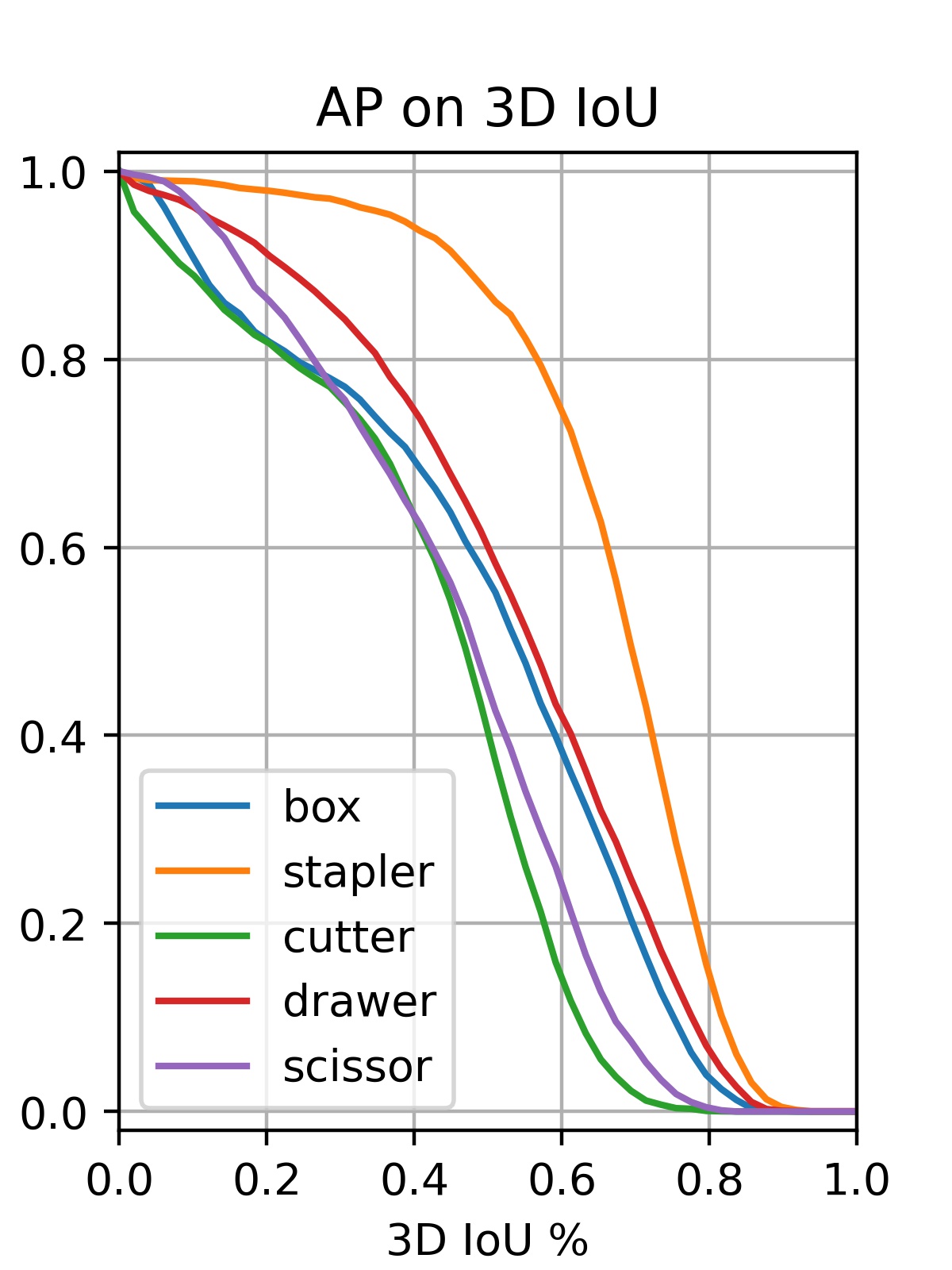}
    \end{subfigure}
    \caption{Articulated pose estimation results on ReArtMix dataset.}
    \label{fig:mixed reality data exp}
    \vspace{-0.5cm}
\end{figure}

\begin{figure}[!tb]
    \centering
    \begin{subfigure}[t]{0.3\linewidth}
    \includegraphics[width=\linewidth]{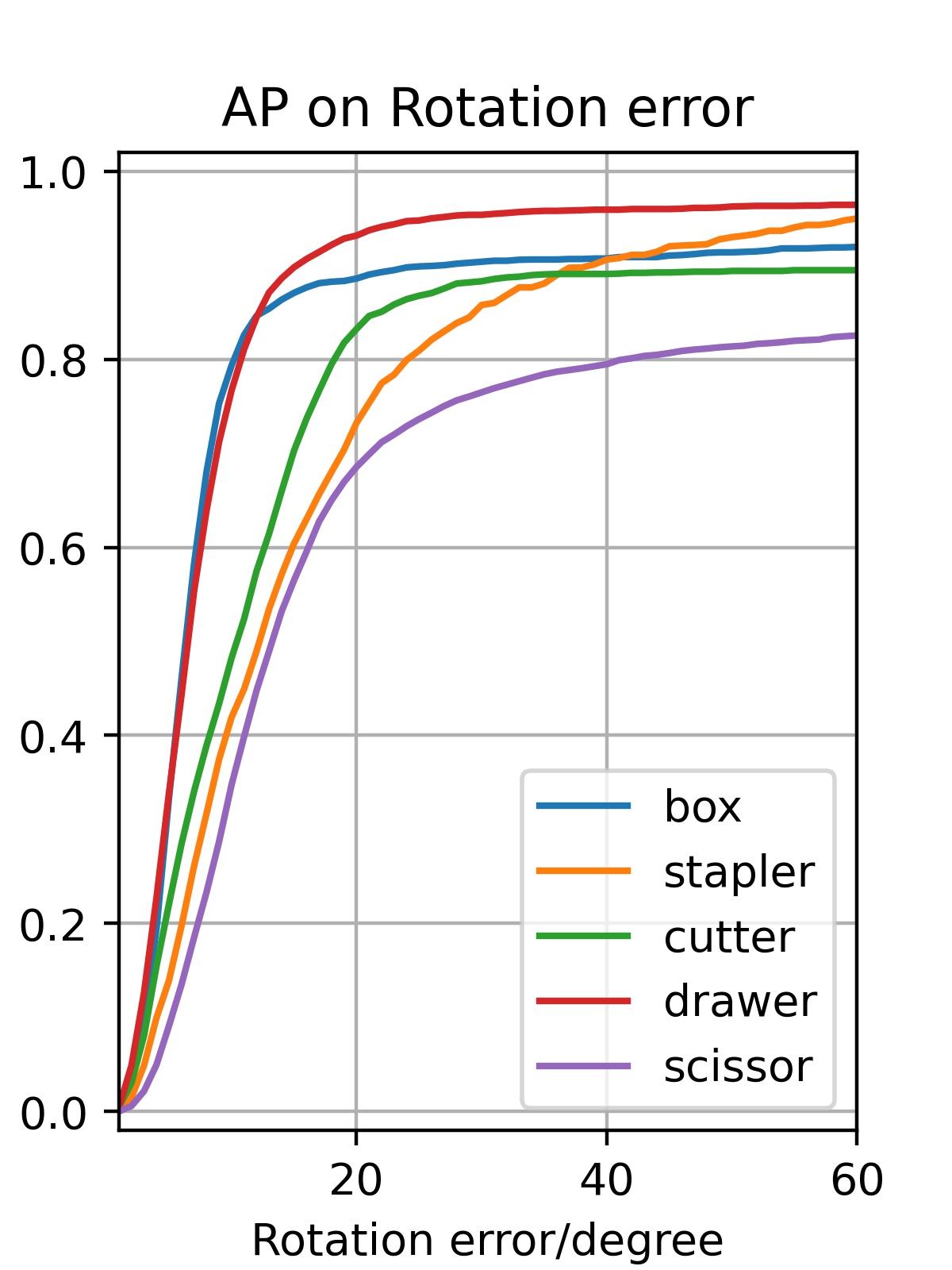}
    \end{subfigure}
    \begin{subfigure}[t]{0.3\linewidth}
    \includegraphics[width=\linewidth]{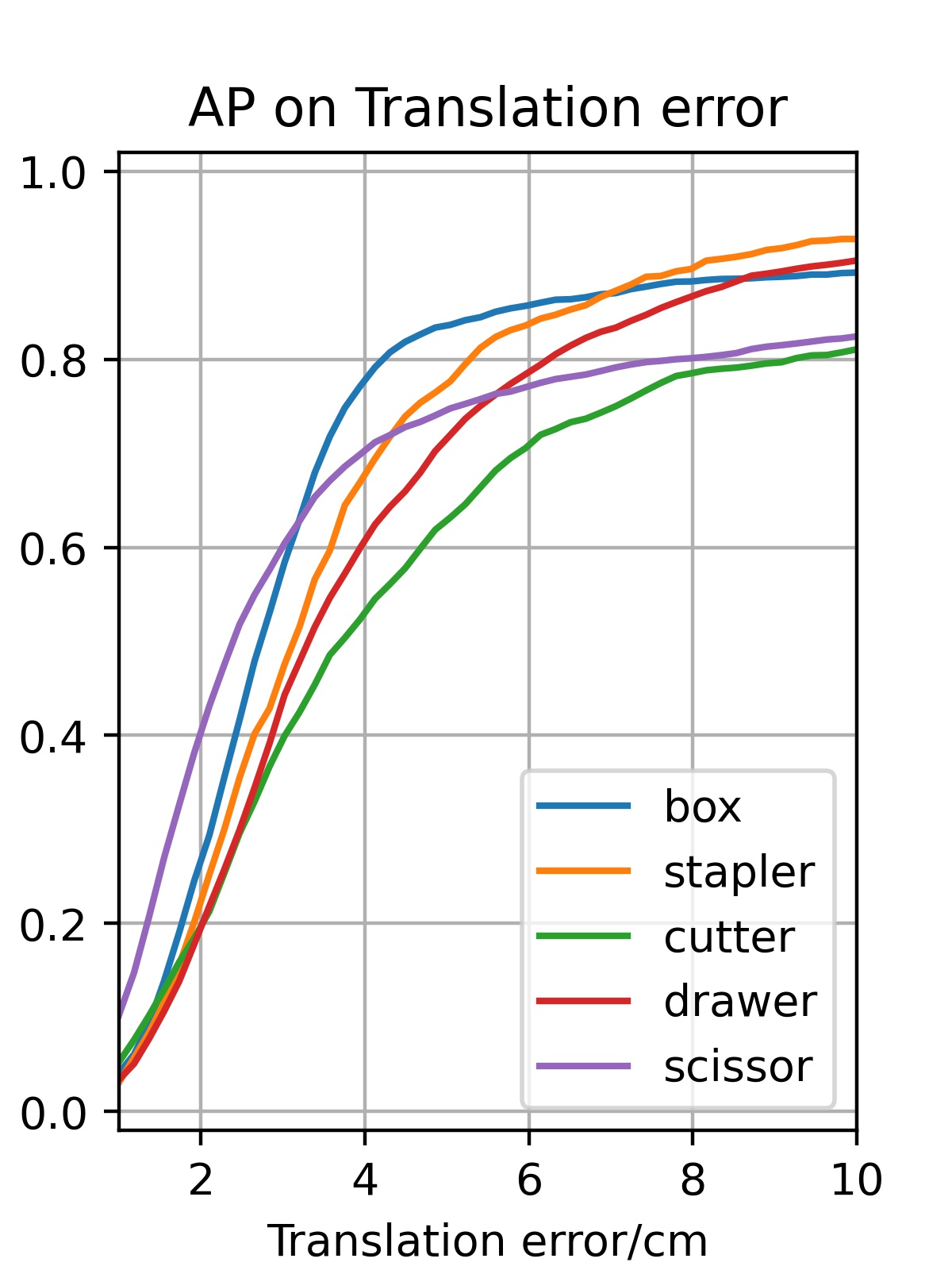}
    \end{subfigure}
    \begin{subfigure}[t]{0.3\linewidth}
    \includegraphics[width=\linewidth]{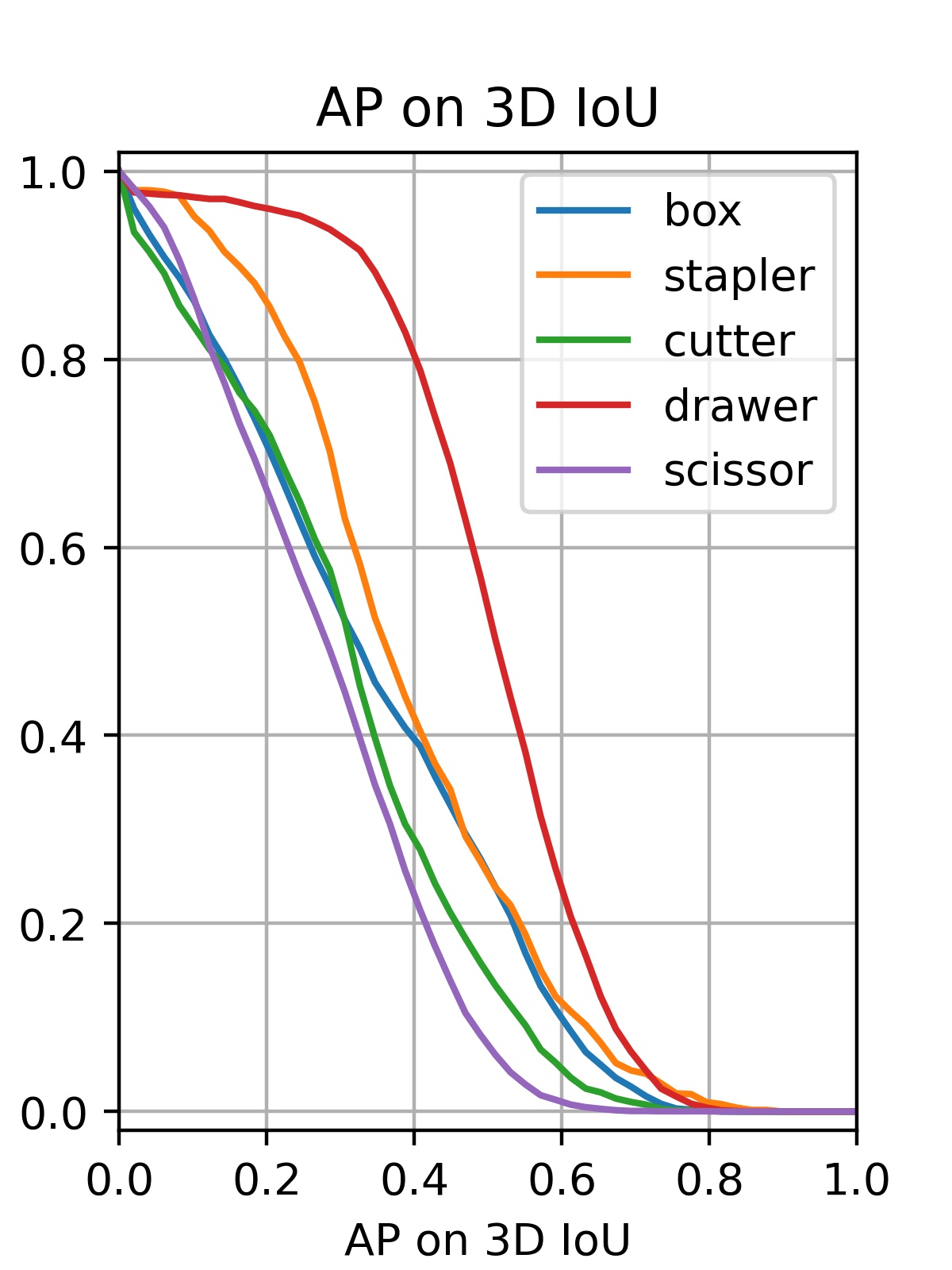}
    \end{subfigure}
    \caption{Articulated pose estimation results on ReArtVal dataset.}
    \label{fig:real world data exp}
    \vspace{-0.5cm}
\end{figure}

\subsection{Implementation Details}

We use RetinaNet \cite{lin2017focal} with ResNet-50 backbone \cite{he2016deep} as our object detector. We use the SGD optimizer with the momentum of 0.9 to train the detector with total training epoch 8. During training PointNet++, we use Adam optimizer with an initial learning rate of 0.001 and 16 batch size. Please refer to supplementary file for more details.

%Dropout 0.5 is adopted in our joint prediction module. 

%train our object detector with ResNet50 backbone with FPN. A detection confidence threshold 0.3 is set to ensure the most objects could be trained and evaluated for pose estimation. 

%\paragraph{Baselines} Since no other methods are designed targeting at CAPER setting, we propose ablated versions of our method to help compare performance. Firstly, we use PointNet++ encoder to extract several feature vectors for all the rigid parts and then we build a joint prediction module that classifies and regress joint information directly from each part descriptor, which is named ReArtNOCS-Reg. The second one is that we compare our method to the version with class-agnostic prediction, in which we predict per-point joint information in the class-agnostic channels.

\paragraph{Metrics} In evaluation with CAPER setting, we report Average Precision (AP) over all the parts of each category, for which the error is less than 5cm and 10cm for translation, 5$^\circ$ and 10$^\circ$ for rotation. We also average the AP over various error thresholds with AP$_{1^\circ:10^\circ}$ for rotation and AP$_{1cm:10cm}$ for translation. In terms of 3D IoU, we use 0.5 and 0.7 as the threshold to report AP and AP$_{0.5:0.7}$. When evaluating with CAPE setting, we use pose accuracy at 10$^\circ$, 10cm, and 3D IoU 0.7 since this setting does not require object detection.

\begin{table}[t]
% \scriptsize
% \centering\small
\small
\centering
\resizebox{\linewidth}{!}{
\begin{tabular}{c|lcc|ccc}
\hline
\multirow{2}{*}{Category} & \multirow{2}{*}{Method} & \multirow{2}{*}{JT} & \multirow{2}{*}{KS} & \multicolumn{3}{c}{Pose Accuracy} \\
\cline{5-7}
 &  & & & 10$^\circ$ & 10cm & 3D$_{70}$ \\
\hline
\multirow{4}{*}{Drawer} & NAOCS & - & gt & 70.6 & 15.0 & - \\
& NPCS & - & gt & 71.9 & 36.8 & 52.4 \\
& A-NCSH & gt & gt & 90.4 & 42.3 & 52.5 \\
& ReArtNOCS & pred & pred & 84.5 & 39.7 & 48.2 \\
\hline
\multirow{4}{*}{Refrigerator} & NAOCS & - & gt & 60.5 & 7.2 & - \\
& NPCS & - & gt & 64.6 & 19.0 & 18.1 \\
& A-NCSH & gt & gt & 69.4 & 21.2 & 20.3 \\
& ReArtNOCS & pred & pred & 65.5 & 18.3 & 14.9 \\
\hline
\multirow{4}{*}{Trashcan} & NAOCS & - & gt & 65.4 & 12.4 & - \\
& NPCS & - & gt & 66.5 & 20.1 & 33.2 \\
& A-NCSH & gt & gt & 77.9 & 24.3 & 33.5 \\
& ReArtNOCS & pred & pred & 69.1 & 22.8 & 31.9 \\
\hline
\end{tabular}}
\caption{Performance comparison on PartNet-Mobility dataset. JT and KS indicate that whether the joint type and kinematic structure are given as ground truth or not.}
\label{tab:sapien_experiment}
\vspace{-0.5cm}
\end{table}

\begin{figure*}[!thb]
    \centering
    \includegraphics[width=\linewidth]{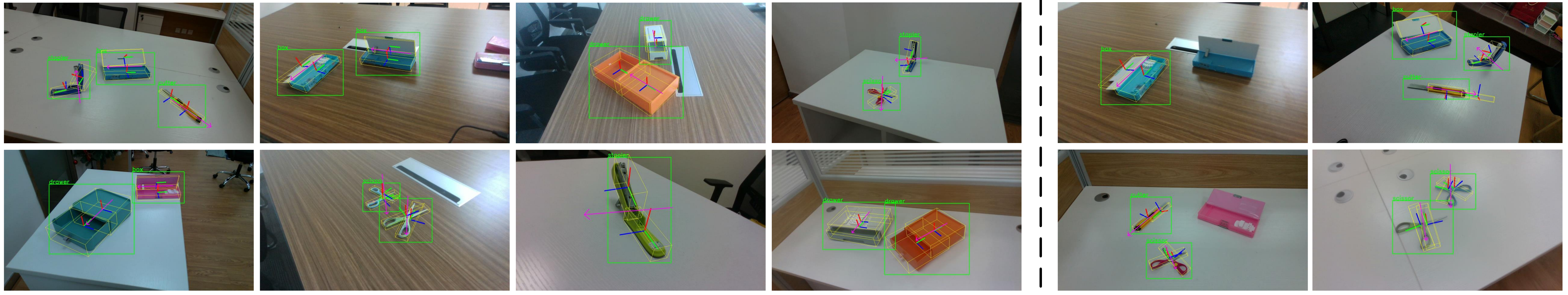}
    \caption{Qualitative results (left) and typical failures (right).}
    \label{fig:qualitative results real data}
    \vspace{-0.2cm}
\end{figure*}

\begin{table*}[!thb]
% \scriptsize
% \centering\small
\tiny
\centering
\resizebox{\linewidth}{!}{
\begin{tabular}{c|c|ccc|ccc|ccc}
\hline
\multirow{2}{*}{Category} & \multirow{2}{*}{Method} & \multicolumn{3}{c}{Rotation Error} & \multicolumn{3}{|c}{Translation Error} & \multicolumn{3}{|c}{3D IoU} \\
\cline{3-11}
 & & AP$_{1:10}$ & AP$_{5}$ & AP$_{10}$ & AP$_{1:10}$ & AP$_{5}$ & AP$_{10}$ & AP$_{0.5:0.7}$ & AP$_{0.5}$ & AP$_{0.7}$ \\
\hline
\multirow{3}{*}{Box} & ReArtNOCS-Reg & 42.6 & 33.4 & 79.3 & 68.9 & 85.7 & 89.2 & 27.6 & 39.8 & 8.2 \\
& ReArtNOCS-ClassAgnostic & \textbf{49.5} & \textbf{46.8} & \textbf{89.2} & \textbf{77.5} & 88.7 & 92.9 & 30.1 & 49.3 & 10.1 \\
& ReArtNOCS & 47.6 & 43.3 & 88.0 & 77.3 & \textbf{90.7} & \textbf{94.1} & \textbf{32.4} & \textbf{52.8} & \textbf{12.5} \\
\hline
\multirow{3}{*}{Stapler} & ReArtNOCS-Reg & 18.9 & 13.9 & 41.9 & 66.3 & 83.6 & 92.8 & 32.7 & 53.7 & 12.2\\
& ReArtNOCS-ClassAgnostic & 18.7 & 14.0 & 41.2 & 67.5 & 79.0 & 92.6 & 39.6 & 60.6 & 19.5 \\
& ReArtNOCS & \textbf{23.3} & \textbf{18.4} & \textbf{47.5} & \textbf{72.9} & \textbf{85.9} & \textbf{95.4} & \textbf{46.6} & \textbf{67.6} & \textbf{26.4} \\
\hline
\multirow{3}{*}{Cutter} & ReArtNOCS-Reg & 24.2 & 22.1 & 48.2 & 56.5 & 70.9 & 81.1 & 10.1 & 24.0 & 4.5 \\
& ReArtNOCS-ClassAgnostic & 28.1 & 27.7 & 51.4 & \textbf{66.6} & 75.5 & 87.4 & 12.8 & 24.3 & 5.2 \\
& ReArtNOCS & \textbf{30.7} & \textbf{29.6} & \textbf{54.0} & 65.6 & \textbf{75.8} & \textbf{87.5} & \textbf{14.4} & \textbf{26.2} & \textbf{8.7} \\
\hline
\multirow{3}{*}{Drawer} & ReArtNOCS-Reg & 38.6 & 33.9 & 76.7 & 62.7 & 73.4 & 86.5 & 48.7 & 66.3 & 21.2\\
& ReArtNOCS-ClassAgnostic & 44.8 & 42.4 & 82.2 & 65.0 & 78.5 & 90.5 & 51.6 & 72.8 & 25.1 \\
& ReArtNOCS & \textbf{48.9} & \textbf{45.6} & \textbf{87.2} & \textbf{71.6} & \textbf{84.1} & \textbf{94.1} & \textbf{62.3} & \textbf{80.4} & \textbf{32.7} \\
\hline
\multirow{3}{*}{Scissor} & ReArtNOCS-Reg & 19.7 & 14.4 & 47.2 & 68.7 & 77.9 & 82.6 & 12.3 & 31.8 & 2.0 \\
& ReArtNOCS-ClassAgnostic & 20.2 & 16.0 & 48.5 & 72.8 & 80.5 & 87.6 & 15.5 & 36.1 & 3.4 \\
& ReArtNOCS & \textbf{29.6} & \textbf{28.1} & \textbf{63.3} & \textbf{76.6} & \textbf{84.9} & \textbf{91.1} & \textbf{18.6} & \textbf{37.7} & \textbf{5.3} \\
\hline
\hline
\multirow{3}{*}{mean} & ReArtNOCS-Reg & 28.8 & 23.5 & 58.6 & 64.6 & 78.3 & 86.4 & 26.3 & 43.1 & 9.6 \\
& ReArtNOCS-ClassAgnostic & 32.3 & 29.4 & 62.5 & 69.9 & 80.4 & 90.2 & 29.9 & 48.6 & 12.6 \\
& ReArtNOCS & \textbf{36.0} & \textbf{33.0} & \textbf{67.9} & \textbf{72.8} & \textbf{84.3} & \textbf{92.4} & \textbf{34.9} & \textbf{52.8} & \textbf{17.1} \\
\hline  
\end{tabular}}
\caption{Performance comparison on ReArtVal dataset. \textbf{ReArtNOCS-Reg} and \textbf{ReArtNOCS-ClassAgnostic} are ablated versions as baselines. \textbf{ReArtNOCS-Reg:} we classify and regress joint properties directly from segmented part features. \textbf{ReArtNOCS-ClassAgnostic:} The ReArtNOCS method adopts class-agnostic strategy to predict joint properties by per point voting.}
%The notion of ReArtNOCS-Reg and ReArtNOCS-Vote w CA please refer to Sec. \ref{sec:ablation}.}
\label{tab:comparison exp}
\vspace{-0.3cm}
\end{table*}

\subsection{Results with CAPER setting}

We report the results of ReArtNOCS training on the ReArtMix training set while testing on the ReArtMix test set and ReArtVal set respectively. In ReArtMix test set, our method achieves mean AP with \textbf{50.6\%}, \textbf{86.7\%} and \textbf{59.7\%} for rotation error 5$^\circ$, translation error 5cm and 3D IoU@0.5. Specifically, ReArtNOCS performs \textbf{77.6\%} AP on rotation error 5$^\circ$ for box, which outperforms other categories. On the contrary, AP on translation for the drawer is much worse than others with only \textbf{63.3\%} for translation error 5cm. This could be explained by the larger average size of drawers in our dataset, which increases the difficulty in estimating their precise locations. See more details in Fig. \ref{fig:mixed reality data exp}.

When testing on ReArtVal set, there appears a drop for all the categories compared to ReArtMix evaluation. In real world data ReArtVal, our method could obtain \textbf{33.0\%}, \textbf{84.3\%} and \textbf{52.8\%} for rotation error 5$^\circ$, translation error 5cm and 3D IoU 0.5. Even though, our ReArtNOCS method could still obtain a comparable performance on drawer and box with only 8.9\% and 7.2\% AP drop for rotation error 10$^\circ$ as well as 3.0\% and 1.2\% drop for translation error 10cm. This indicates that ReArtNOCS could partly address pose estimation issue in real world. See more details on Fig. \ref{fig:real world data exp} and Table \ref{tab:comparison exp}.

\subsection{Results with CAPE setting}

We also evaluate our method on a public synthetic articulated model repository PartNet-Mobility \cite{xiang2020sapien} for the CAPE task. We select 91 models from three categories, including drawer, refrigerator, and trashcan with various kinematic structures. We follow the rendering pipeline in \cite{li2020category} to generate synthetic images with these models. For each category, we have 5,000 images for training and 1,000 images for testing.  

In Table \ref{tab:sapien_experiment}, we report the performance of our ReArtNOCS along with some strong baselines, namely NAOCS, NPCS and A-NCSH \cite{li2020category}, in which A-NCSH uses joint properties as ground truth so it performs as the upper bound of our method. Comparing to NAOCS and NPCS methods, ReArtNOCS shows better performance that achieves average \textbf{71.3\%}, \textbf{27.0\%} and \textbf{31.6\%} accuracy for 10$^\circ$, 10cm and 3D IoU of 0.7. For A-NCSH method, we could also obtain comparable result on drawer and refrigerator, where there are only \textbf{5.9\%}, \textbf{3.9\%} and \textbf{8.8\%} accuracy disparity on rotation error 10$^\circ$.

\subsection{Ablation Study}\label{sec:ablation}
We conduct ablation study on the mechanism to estimate the joint property. Please refer to supplementary file for more ablation studies.

\textbf{Reg. vs. Point Vote} We compare the results using per point voting strategy with directly regression strategy. On the ReArtVal dataset, the dense prediction per-point/pixel method is consistently better than direct regression in the first two rows in Table \ref{tab:comparison exp}. Because direct regression using global part feature which largely relies on the precise part segmentation accuracy.

\textbf{Class-Agnostic vs. Class-Aware} The class-aware prediction could obviously bring a superior performance than using class-agnostic strategy since it predicts joint properties in separate channels for different parts, even though it might cause higher computational costs in our network. As it can be seen in Table \ref{tab:comparison exp}, there appears \textbf{3.7\%} and \textbf{2.9\%} AP improvement for CAPER task. It also benefits the performance on AP with 3D IoU, which achieves 5.0\% AP improvement on AP$_{0.5:0.7}$.

\subsection{Qualitative Results}

Qualitative results on ReArtVal are displayed in Fig. \ref{fig:qualitative results real data}. We also summarize the failure of the CAPER task into the following reasons: (1) Detection missing. There exist a domain gap between mixed reality data and real-world data that influence detection accuracy. (2) Quality of depth image. Depth camera holds its limitation on inaccurate depth map, especially when the scissor or cutter lays flat on the table.

%We visualize some of the qualitative results in Fig. \ref{fig:qualitative results real data}. In addition, we summarize the failure of the CAPER task into the following reasons: (1) Detection missing. There exist a domain gap between mixed reality data and real-world data that influence detection accuracy. (2) Quality of depth image. Depth camera holds its limitation on inaccurate depth map captured, especially the scissor or cutter lay flat on the table.

\begin{comment}
\begin{figure*}[htb]
    \centering
    \begin{subfigure}[t]{0.66\linewidth}
    \includegraphics[width=\linewidth]{figures/Qualitative Results real2.png}
    \caption{Qualitative Results}
    \end{subfigure}
    \begin{subfigure}[t]{0.33\linewidth}
    \includegraphics[width=\linewidth]{figures/Qualitative Results failure cases.png}
    \caption{Failure cases}
    \end{subfigure}
    \caption{Qualitative results (left) and typical failures (right).}
    \label{fig:qualitative results real data}
\end{figure*}
\end{comment}

\section{Conclusion}

In this paper, we extend the CAPE task and formulate the CAPER problem for real-world articulation pose estimation. Accompanying the task setting, we provide a full package of solutions including the FAOM-SAMERT pipeline to semi-automatically build the dataset for the CAPER, the effective framework ReArtNOCS that could deal with various kinematic structures, and multiple-instance occurrence issues. We hope the proposed CAPER task can help the researchers to rethink the CAPE task setting, and the proposed dataset generation pipeline and learning framework can serve as a strong baseline for future research.

%We formulate a novel challenge part-level 6D pose estimation for previously unseen articulated instances with unknown kinematic structure. In dealing with this problem, we build our domain-specific benchmark dataset consisting of mixed reality data and real world data, in which the dataset is built upon a new articulated model repository. Accompanying with the problem and dataset, we propose an effective framework containing 2D articulated object detector and 3D pose estimator. To cope with the unseen instances with unknown kinematic structure, we present a joint prediction module based on PointNet++ architecture that can be used with the 3D local point cloud and part-level NOCS map to estimate the full metric 6D pose per part. Our approach shows potential value on practical applications such as robotics and 3D scene understanding.

\newpage
\bibliographystyle{named}
\bibliography{ms}

\end{document}